\documentclass[preprint,1p]{elsarticle}

\usepackage{amsmath}
\usepackage{amssymb}
\usepackage{hyperref}
\usepackage{todonotes} 
\usepackage{amsmath}
\usepackage{soul}
\usepackage{algorithm}
\usepackage{algorithmic}
\usepackage{amsthm}
\usepackage{booktabs}
\usepackage{graphicx}

\theoremstyle{plain}
\newtheorem{theorem}{Theorem}

\begin{document}
\journal{UNSURE}
\title{PEARL: Preconditioner Enhancement through Actor-critic Reinforcement Learning}

\author[1]{David Millard}
\ead{djm3622@rit.edu}

\author[2]{Arielle Carr}
\ead{arg318@lehigh.edu}

\author[3]{Stéphane Gaudreault}
\ead{stephane.gaudreault@ec.gc.ca}

\author[4]{Ali Baheri\corref{cor1}}
\ead{akbeme@rit.edu}

\cortext[cor1]{Corresponding author}

\address[1]{Golisano College of CIS, Rochester Institute of Technology, Rochester, USA}
\address[2]{Computer Science \& Engineering, Lehigh University, Bethlehem, USA}
\address[3]{Recherche en prévision numérique atmosphérique, Environnement et Changement climatique Canada, Dorval, Canada}
\address[4]{Kate Gleason College of Engineering, Rochester Institute of Technology, Rochester, USA}




\begin{abstract}
\sloppy
 We present PEARL (\textbf{\ul{P}}reconditioner \textbf{\ul{E}}nhancement through \textbf{\ul{A}}ctor-critic \textbf{\ul{R}}einforcement \textbf{\ul{L}}earning), a novel approach to learning matrix preconditioners. Existing preconditioners such as Jacobi, Incomplete LU, and Algebraic Multigrid methods offer problem-specific advantages but rely heavily on hyperparameter tuning. Recent advances have explored using deep neural networks to learn preconditioners, though challenges such as misbehaved objective functions and costly training procedures remain. PEARL introduces a reinforcement learning approach for learning preconditioners, specifically, a contextual bandit formulation. The framework uses an actor-critic model, where the actor generates the incomplete Cholesky decomposition of preconditioners, and the critic evaluates them based on reward-specific feedback. To further guide the training, we design a dual-objective function, combining updates from the critic and condition number. PEARL contributes a generalizable preconditioner learning method, dynamic sparsity exploration, and cosine schedulers for improved stability and exploratory power. We compare our approach to traditional and neural preconditioners, demonstrating improved flexibility and iterative solving speed.
\end{abstract}

\maketitle


\section{Introduction}\label{sec:introduction}
Many problems involving complex partial differential equations (PDEs) result in the discretization of space, producing large systems of linear equations. Solving these systems often requires advanced iterative solvers due to their size and complexity. Iterative solvers are foundational tools used in tasks such as numerical weather prediction, power system analysis, and fluid dynamics simulations. The behavior of these solvers is well-supported by a rich body of literature, much of which focuses on optimization techniques. This papers focuses is on the well-established domain of preconditioning \cite{https://doi.org/10.1002/nla.716}.

A preconditioner is a matrix or operator that transforms a system of linear equations to facilitate faster or more efficient solutions using iterative methods \cite{doi:10.1137/1.9780898718003}. Existing preconditioners include the Jacobi method, incomplete LU factorization (ILU), incomplete Cholesky factorization, and algebraic multigrid (AMG), each offering distinct advantages depending on the problem's characteristics \cite{BENZI2002418}. Recent advancements have largely been attributed to the intelligent adaption of deep neural networks to approximate effective preconditioners. A preconditioner's measure of success is straightforward: the speedup achieved. This is typically evaluated using two key metrics—solving time and the number of iterations required for convergence. A third desirable property is ease of construction.

The primary challenge in recruiting deep neural networks to approximate effective preconditioners lies within the training procedure. Striking a balance between reasonable and efficacious objective functions is delicate, particularly because the two main objectives of success—minimizing solving time and the number of iterations—do not have gradients accessible by the model. Another significant challenge is the construction time of the preconditioner. Existing preconditioning techniques are computationally inexpensive and can be constructed quickly, but their effectiveness depends heavily on hyperparameter selection. In contrast, deep learning techniques are more general-purpose, but they require an extremely costly training procedure. Therefore, it is essential to design these methods in the context of learning mappings for specific classes of systems.

Our approach tackles these challenges by introducing a novel method that generalizes beyond previous work, giving the model greater flexibility and control in the discovery of effective preconditioners. This is achieved by indirectly minimizing the number of iterations needed for convergence. To this end, we frame the challenge as a reinforcement learning problem, specifically as a contextual bandit problem \cite{9185782}. In this framework, the state space is defined by the linear system, and the action space consists of the predicted optimal preconditioners \cite{duckworth2023reinforcement}. The reward is based on the number of iterations required for convergence. Using an actor-critic \cite{NIPS1999_6449f44a} framework, we aim to effectively navigate the complex problem space, allowing the model greater exploration within the preconditioner space.

\noindent {\textbf{Our Contributions.}} The contributions of this paper are threefold:

\begin{itemize}
    \item We develop a novel actor-critic reinforcement learning framework for learning matrix preconditioners, introducing a contextual bandit formulation that enables direct optimization of solver performance rather than relying solely on condition number metrics.
    \item We establish theoretical guarantees for convergence rates and prove improved complexity bounds, bridging the gap between practical performance and theoretical understanding.
     \item We demonstrate that our dual-objective optimization strategy with cosine scheduling achieves solver speedup while enabling broader exploration of preconditioner structures, verified through both empirical convergence results and visualization of learned preconditioner patterns.
\end{itemize}

\noindent {\textbf{Paper Organization.}} The remainder of this paper is structured as follows. Section 2 reviews literature in preconditioner learning, while Section 3 establishes foundational concepts in conjugate gradient methods and actor-critic reinforcement learning. Section 4 presents our PEARL methodology, detailing the training data generation, model architectures, and cosine scheduling mechanism. Section 5 provides theoretical analysis of convergence rates and complexity bounds, followed by Section 6 which presents numerical experiments. The paper concludes with discussion of key findings in Section 7 and future research directions in Section 8.

\section{Literature Review}
Recent works use constrained optimization for three core techniques: learning to approximate the non-zero entries of traditional methods \cite{häusner2024learningincompletefactorizationpreconditioners}, optimizing for speedups of traditional methods \cite{luz2020learningalgebraicmultigridusing}, and predicting direct inverse preconditioners \cite{sappl2019deeplearningpreconditionersconjugate}. Other approaches use fourier neural operators (FNOs) \cite{li2021fourierneuraloperatorparametric} and physics-informed neural networks (PINNs)\cite{MAO2020112789}, can further constrain the system based on the properties of the PDEs. Our approach aims to develop a general-purpose algorithm, similar to Chen's work \cite{chen2024graphneuralpreconditionersiterative}, for directly predicting inverse preconditioners. We extend this large body of work by employing reinforcement learning, allowing feedback from iterative solvers to indirectly backpropagate through the model during training \cite{9545961}.

While our formulation is unique to this field, it is not without precedent. A similar challenge is faced in recommender systems, where the response to a targeted ad depends on user feedback. Since this feedback is not a direct transformation of the input, it provides no gradient to correct errors generated by the model \cite{9185782}. In our approach, we treat the weighted sum of convergence iterations, the percentage of zeros, and the magnitude of the final iteration as feedback signals from the iterative solver. However, rather than relying solely on this weighted sum, we also minimize the condition number, which is differentiable but numerically unstable. Using a dual optimization approach, we aim to enhance stability under the guidance of both objective function. We believe this enriches  the model with more information, enabling it to maximize its exploration of the preconditioner space.

Literature has increasingly favored the use of graph neural networks (GNNs) \cite{chen2024graphneuralpreconditionersiterative, luz2020learningalgebraicmultigridusing, häusner2024learningincompletefactorizationpreconditioners} due to their unique balance between scalability and effective information translation \cite{10.1007/978-3-030-99372-6_7}. When discretizing PDEs, the size of the resulting matrices are exceptionally large. While Convolutional Neural Networks (CNNs) are known for their scalability, their performance often suffers in the presence of extreme sparsity \cite{8374553}. On the other hand, fully connected linear models excel in information translation but struggle with higher dimensional spaces \cite{millard2024deeplearningkoopmanoperator}. GNNs, although largely favored for their blend of both scalability and translation, suffer from fixed sparsity patterns. In our formulation, we prioritize the flexibility to learn sparsity patterns dynamically. As a result, we have chosen fully connected networks that operate on low-dimensional data.

The challenge of balancing exploration with stability constraints shares conceptual similarities with safe reinforcement learning approaches \cite{garcia2015comprehensive,yifru2024concurrent,baheri2020deep,baheri2022safe}. In safe RL, the agent must explore the action space while respecting safety specifications. Similarly, in preconditioner learning, we must explore the space of possible preconditioners while maintaining numerical stability through condition number constraints. Our work builds on these insights from safe RL by adopting a dual-objective framework that explicitly manages this exploration-stability tradeoff.

\section{Preliminaries}

\subsection{Conjugate Gradient Method}

The conjugate gradient (CG) \cite{https://doi.org/10.1002/wics.13} method is an iterative algorithm for solving systems of linear equations of the form: 
\begin{equation}
\mathbf{A} \mathbf{x} = \mathbf{b},
\end{equation}
where \(\mathbf{A}\) is a symmetric positive-definite matrix, \(\mathbf{x}\) is the vector of unknowns, and \(\mathbf{b}\) is a known vector. This method is particularly efficient for sparse matrices, such as those arising from the discretization of partial differential equations (PDEs). The CG method minimizes the quadratic form 
\begin{equation}
Q(\mathbf{x}) = \frac{1}{2} \mathbf{x}^\top \mathbf{A} \mathbf{x} - \mathbf{b}^\top \mathbf{x},
\end{equation}
by iteratively updating the solution vector \(\mathbf{x}\) along mutually conjugate directions. These directions ensure that each iteration reduces the error optimally in the subspace spanned by all previous directions, leading to faster convergence compared to basic gradient descent. This method is designed to converge within $n$ iterations, where $n$ represents the dimensions of the matrix. If the method fails to converge within this limit, it is generally considered intractable \cite{doi:10.1137/1.9780898718003}.

\subsubsection{Preconditioning the Conjugate Gradient Method}

For ill-conditioned systems, where the condition number of \(\mathbf{A}\) is large, the convergence of the CG method can be significantly slowed. To address this, a preconditioner \(\mathbf{M}\) is introduced, transforming the system into an equivalent but better-conditioned form:
\begin{equation}
\mathbf{M}^{-1} \mathbf{A} \mathbf{x} = \mathbf{M}^{-1} \mathbf{b}.
\end{equation}
The matrix \(\mathbf{M}\) should approximate \(\mathbf{A}\) but be easier to invert or apply. The choice of \(\mathbf{M}\) aims to reduce the condition number of \(\mathbf{M}^{-1} \mathbf{A}\), thereby improving convergence. The preconditioned CG method often results in dramatic reductions in the number of iterations required for convergence, especially for systems arising from discretized PDEs or other ill-conditioned problems.

\subsection{Contextual Bandit}

A contextual bandit is a decision-making problem where an agent selects an action based on an observed context and receives a reward for that action \cite{9185782}. The goal is to maximize cumulative rewards by learning which actions are optimal in different contexts. At each timestep \(t\), the agent observes a context \(\mathbf{c}_t \in \mathcal{C}\), selects an action \(a_t \in \mathcal{A}\), and receives a reward \(r_t \sim \mathcal{R}(\mathbf{c}_t, a_t)\). Unlike standard reinforcement learning, contextual bandits do not account for long-term consequences, instead focusing on immediate rewards.

The goal of the contextual bandit algorithm is to learn a policy \(\pi(a | \mathbf{c})\) that maps contexts to actions in a way that maximizes the expected cumulative reward over time:
\begin{equation}
\pi^* = \arg \max_\pi \mathbb{E}\left[\sum_{t=1}^T r_t\right].    
\end{equation}
 Our approach extends this formulation, enabling the model to prioritize the primary objective of minimizing convergence iterations while simultaneously addressing secondary goals in a balanced manner. We adapt this paradigm to our problem by defining the context as the matrix $(\mathbf{A} + \Delta \mathbf{A})$ from equation (\ref{eq:dataset}), and the action as the inverse preconditioner, a matrix with the same dimensions. Our reward is denoted as as a linear combination of key metrics defined in depth by equation (\ref{eq:reward}).

\subsection{Actor-Critic}

Actor-critic methods are a class of reinforcement learning algorithms that combine value-based and policy-based models together \cite{NIPS1999_6449f44a}. These methods consist of two components: the actor and the critic. The actor is responsible for exploring the action space and updating the policy, while the critic provides feedback by estimating how good the selected actions are. The policy is updated using the gradient of the expected reward:
\begin{equation}
\nabla_\theta J(\theta) = \mathbb{E}_{s \sim d^\pi, a \sim \pi_\theta} \left[ \nabla_\theta \log \pi_\theta(a|s) \, Q(s, a) \right],
\end{equation}
where \(Q(s, a)\) is the action-value function. In the context of a contextual bandit problem, the state space is defined by a one-step MDP and the critic estimates \(Q(s, a)\) as the immediate reward $r_t$. The critic’s parameters \(w\) are updated to minimize this error:
\begin{equation}
w \gets w - \alpha \nabla_w \delta_t^2.
\end{equation}
Actor-critic methods are advantageous for their ability to handle continuous action spaces and to directly learn stochastic or deterministic policies. However, they often suffer from instability due to the interdependence of the actor and critic updates, which can lead to divergence if not carefully managed \cite{osinenko2022notestabilizingreinforcementlearning}. Stabilizing these updates typically involves techniques such as gradient clipping or imposing restrictions to prevent abrupt changes during the training procedure.

In our formulation, the actor is responsible for outputting a preconditioner matrix $\mathbf{M}$, which is intended to transform the original system into one that is better conditioned. The critic evaluates the effectiveness of the proposed preconditioner by computing a reward that reflects the performance improvement. This reward is based on a combination of factors, such as the reduction in iteration count for solving the system, magnitude of the final residual, and sparsity of the preconditioner.

\section{Methodology}

\subsection{Training Data Generation}

We generate our dataset by discretizing 2D diffusion equations \cite{https://doi.org/10.1002/fld.1650090107}, which models the distribution of a heat across a two-dimensional plane over time. The equation is given by: 
\begin{equation}
-\nabla \cdot \left( a(x, y) \nabla \psi \right) = f(x, y),
\end{equation}
where \(a(x, y)\) is the spatially varying diffusion coefficient, \(\psi\) is the unknown function, and \(f(x, y)\) is a source term. To construct the coefficient \(a(x, y)\), we define:
\begin{equation}\label{eq:diff}
a(x, y) = 1 + \sum_{i=1}^{N_b} h_i \exp\left(-\frac{(x - x_i)^2 + (y - y_i)^2}{2w_i^2}\right),
\end{equation}
where \(N_b\) is the number of Gaussian bumps, \(h_i\) is the amplitude of the \(i\)-th bump, \((x_i, y_i)\) is its center, and \(w_i\) is its width. We then discretize the diffusion operator over an \(n \times n\) uniform grid with grid spacing \(h\).

To create the dataset, we first initialize equation (\ref{eq:diff}) with random values sampled from $N_b \sim U(3, 10)$ and $h_i \sim U(0.1, 5)$. The equation is then discretized on an $n \times n$ grid \cite{doi:10.1137/S0036142992232949}, producing a matrix $\mathbf{A}$ with non-zero entries on the main and off-diagonal. We then introduce a random perturbation onto the nonzero entries of $\mathbf{A}$. Finally, we sample an arbitrary vector $\mathbf{b}$ from a multivariate normal distribution of size $n \times 1$. The final formulation of the dataset can be represented as:
\begin{equation}\label{eq:dataset}
    (\mathbf{A} + \mathbf{\Delta A}) \mathbf{x} = \mathbf{b},
\end{equation}
where $\mathbf{A}$ is the discretized 2D diffusion matrix with  $N_b \sim U(3, 10)$ and $h_i \sim U(0.1, 5)$, and $\mathbf{\Delta A}$ is a random perturbation, defined such that $\mathbf{\Delta A}_{i,j} = 0 \text{ for all } \mathbf{A}_{i,j} = 0$ and $\mathbf{\Delta A}_{i,j} \sim U(0, m) \text{ for all } \mathbf{A}_{i,j} \not= 0$. The right-hand side vector $\mathbf{b}$ is drawn from a normal distribution $\mathbf{b} \sim N(\mu, \Sigma)$, and $\mathbf{x}$ is some unknown vector.

These resulting systems are symmetric and positive definite. Their most prominent feature is extreme ill-conditioning. The 2D diffusion equation, particularly for large values of $h_i$, is notorious for producing highly ill-conditioned matrices. When combined with random perturbations, empirical results show that solving these systems, utilizing no preconditioner, becomes intractable. Although the random perturbations detract from the model's physical interpretability, they introduce additional complexity and facilitate the mapping of a class of ill-conditioned systems to a set of effective preconditioning techniques. Our experiments use a \( 25 \times 25 \) discretization, resulting in condition numbers typically ranging from 3,000 to 12,000.

\subsection{Objectives}
Both our state and action spaces lie in \(\mathbb{R}^{n \times n}\). To handle this, we employ an actor-critic algorithm, which is well-suited for continuous action spaces \cite{NIPS1999_6449f44a}. The actor maps the state space to the optimal action according to the policy network, while the critic estimates the immediate reward. We consider our actor parameterized by $\theta$ and our critic parameterized by $\phi$. Drawing inspiration from \cite{häusner2024learningincompletefactorizationpreconditioners}, we plan to pretrain the actor network in a fully supervised manner, with our loss defined as:
\begin{equation}\label{eq:pretrain_actor}
    \mathcal{L}_{\text{pretrain}} = \frac{\sigma_{\text{max}}}{\sigma_{\text{min}}}  
\end{equation}
where $\sigma_i$ refers to the $i$-th singular value computed from $s_i \cdot \pi_\theta(s_i)$. The primary purpose of this phase is to gather enough data to pretrain the critic across multiple agents. 

During the primary training phase, the critic network minimizes the distance from the reward function, and the actor maximizes the estimated reward. The critic’s objective function is defined as:
\begin{equation}\label{eq:critic_loss}
    \mathcal{L}_\phi = \left(Q_\phi(s_i, a_i) - r\right)^2,
\end{equation}
where \( s_i \) is the sampled state, \( a_i = \pi_\theta(s_i) \) is the action generated by the actor network, \( Q_\phi(\cdot) \) is the critic network’s estimate of the action value function, and \( r \) is the immediate reward received. This formulation is standard in contextual bandit problems. The reward function is defined as:
\begin{equation}\label{eq:reward}
    r = 
        w_1 \cdot \frac{1}{N} + w_2 \cdot \frac{n \cdot n - \|a_i\|_0}{n \cdot n} + w_3 \cdot \|r_n\|_2,
\end{equation}
where \( N \) is the maximum number of iterations allowed, \( n \) is the dimension of the state matrix \( s_i \), \( r_n \) is the residual vector after the \( n \)-th iteration, and \( n \) represents either the iteration at convergence or \( N \) if convergence is not achieved.
The actor’s objective function is:
\begin{equation}\label{eq:actor_loss}
    \mathcal{L}_{\theta} = 
     \gamma\log
        \left((
            \log(\sigma_\text{max}) - \log(\sigma_\text{min})
        \right) - Q_\phi(s_i, a_i),
\end{equation}
where \( \sigma_i \) denotes the \( i \)-th singular value of \( s_i \cdot a_i \), and \( \gamma \) is a scaling weight. The log transformation is applied to align the range of the critic’s estimate with the condition number.

\subsection{Models}

In this paper, we constrain our actor to estimate the incomplete Cholesky decomposition, as it is essential to produce symmetric positive definite (SPD) matrices for effectively preconditioning the CG method \cite{6710599}. Additionally, we enforce the preconditioner estimates to lie within the interval \([0, 1]\) by applying a sigmoid activation function to the final output, defined as
\(
\mathbf{M}_{ij} = \sigma(\mathbf{A}_{ij}),
\)
where \(\mathbf{M}\) is the preconditioner matrix, \(\mathbf{A}\) is the actor's raw output, and $\sigma(\cdot)$ is the sigmoid function \cite{NARAYAN199769}. We find that preconditioners constrained to elements within this range exhibit slower convergence compared to their logit counterparts, but sufficiently reduces the search space, enabling the discovery of a more diverse set of preconditioners within a reasonable experimentation timeframe. Our testing strategy is then divided between two critic models: one that provides a single estimate of the combined reward and another that separates the estimated reward components into multiple outputs. Additionally, we include test cases for the incomplete LU decomposition \cite{https://doi.org/10.1002/nla.1680020208}, as well as various non-symmetric actors, which are detailed in \ref{appendix:actors}\footnote{The codebase for this work, including implementation details and test cases, is publicly available at \url{https://github.com/djm3622/precondition-discovery-contextual-bandit}.}.

\subsubsection{Actor (Incomplete Cholesky Decomposition)}

The actor model, $\pi_\theta$, generates preconditioner matrices $\mathbf{M}$ by computing their incomplete Cholesky decomposition. Given an input system matrix $\mathbf{A} \in \mathbb{R}^{n \times n}$, the model predicts a lower triangular matrix $\mathbf{L} \in \mathbb{R}^{n \times n}$ such that:
\begin{equation}
    \mathbf{M} = \mathbf{L} \mathbf{L}^\top.
\end{equation}
The model is parameterized as a fully connected neural network with variable hidden sizes and number of layers. The network's output represents the lower triangular elements of $\mathbf{L}$, specifically the indices given by:
\begin{equation}\label{eq:param}
\text{tril}(\mathbf{L}) = \{\mathbf{L}_{ij} \mid i \geq j\}.
\end{equation}
To enforce sparsity, the elements of $\mathbf{M}$ are thresholded element-wise based on a sparsity tolerance $\epsilon_\text{tol}$:
\begin{equation}
\mathbf{M}_{ij} = 
\begin{cases} \label{eq:sparse}
\mathbf{M}_{ij}, & \text{if } |\mathbf{M}_{ij}| \geq \epsilon_\text{tol}, \\
0, & \text{otherwise}.
\end{cases}
\end{equation}\label{eq:positive}
Additionally, a diagonal bias term $\alpha \mathbf{I}$ is added for numerical stability, where $\alpha \in \mathbb{R}$ and $\mathbf{I}$ is the identity matrix:
\begin{equation}
\mathbf{M} = \mathbf{L} \mathbf{L}^\top + \alpha \mathbf{I}.
\end{equation}
This ensures positive definiteness in the beginning of the training cycle and can be removed after sufficient training time. The bias term depends on the range of the preconditioner output; however, we found that using \(\alpha = 0.1\) within the constrained range \([0, 1]\) yielded the best results.

\subsubsection{Critic (Single-Reward)}

The single-critic model predicts a scalar reward $r \in \mathbb{R}$ which estimates the reward received from the preconditioner matrix, as defined in \ref{eq:reward}. The model maps the concatenated input $[\mathbf{A}, \mathbf{M}] \in \mathbb{R}^{2n \times n}$ to a scalar output:
\begin{equation}
r = Q_\phi(\mathbf{A}, \mathbf{M}),
\end{equation}
where $Q_\phi$ is a fully connected network with ReLU activations and a linear output layer.

\subsubsection{Critic (Multi-Reward)}

The multi-reward critic model extends the single-reward approach by predicting a vector of rewards $\mathbf{r} \in \mathbb{R}^3$, where:
\begin{equation}
r_1 \approx \frac{1}{N}, \quad r_2 \approx \frac{n^2 - \|\mathbf{M}\|_0}{n^2}, \quad \text{and} \quad r_3 \approx \|\mathbf{r}_n\|_2.
\end{equation}
The model maps the concatenated input $[\mathbf{A}, \mathbf{M}] \in \mathbb{R}^{2n \times n}$ to the reward vector:
\begin{equation}
\mathbf{r} = Q_\phi(\mathbf{A}, \mathbf{M}),
\end{equation}
where $Q_\phi$ is a shared fully connected network with a final multi-output linear layer. Each output corresponds to one of the rewards $r_1$, $r_2$, or $r_3$.

\subsection{Cosine Scheduler}

We introduce a cosine scheduler to control the exploration and conditioning effect throughout training. This approach effectively manages the buffer, which is constrained by the limited available memory due to the size of the matrices being stored. In addition, the cosine scheduler helps balance the dual optimization objectives of the actor model. Specifically, it prevents the condition number minimization from dominating, a common issue since its backpropagation often becomes the primary focus, causing the critic estimates to collapse to zero. This collapse is expected because, although the condition number is numerically unstable, its differential is consistent. Conversely, the critic model, particularly in the early stages of training, exhibits inconsistencies with rapid fluctuations between instances. Despite this, we observe that the scheduler improves performance especially when condition number minimization alone yields good results, as demonstrated later.

For the iteration component of the loss to provide meaningful signals, particularly for systems that are intractable, the condition number can help jump-start the algorithm, provided it is well-behaved. This ensures that the buffer accumulates sufficient experience to support all reward components effectively. Furthermore, the exploration scheduler proves highly effective in training actors without condition number minimization (i.e., in a standalone actor-critic framework), as illustrated by the incomplete LU factorization model output shown in \ref{appendix:C}.

The cosine scheduler, applied to both the conditioning effect and exploration, is parameterized as follows:
\begin{equation}
    v(t) = v_{\text{min}} + \frac{1}{2} \left(1 + \cos\left(2\pi \frac{t}{T} + \phi\right)\right) \cdot (v_{\text{max}} - v_{\text{min}}),
\end{equation}
where $t$ is the current timestep, $T$ is the number of timesteps that make up one period, $v_{\text{min}}$ is minimum value of the scheduler, $v_{\text{max}}$ is maximum value of the scheduler, and $\phi$ is a phase shift applied cosine function.

\begin{figure}
    \centering
    \includegraphics[width=0.7\linewidth]{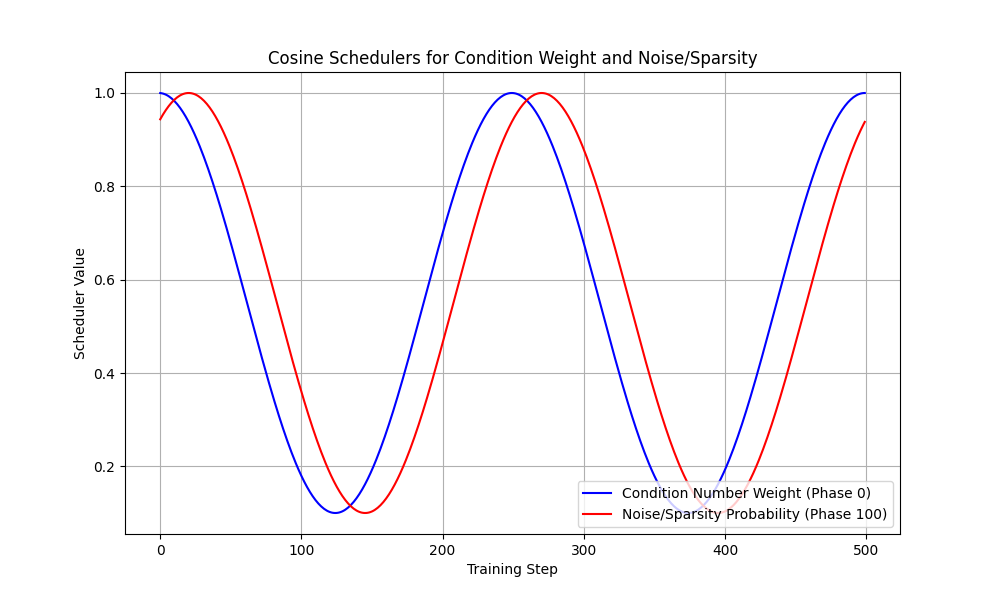}
    \caption{Cosine-scheduler weights are plotted over time, with the waves slightly shifted to introduce additional stochasticity to the crest and troughs of the amplitude.}
    \label{fig:cos-sch}
\end{figure}

\subsection{Training Procedure}

To train both the actor and the critic, we use an experience replay mechanism. We first generate perturbed systems and then apply two forms of exploration noise: symmetric and sparse. Symmetric noise is defined as standard noise sampled from \( N(0, \Sigma) \) with a size of \( \mathbb{R}^{n \times n} \). The noise is applied as follows:
\begin{equation}
M \leftarrow M + (E + E^\top),
\end{equation}
where \( E \) denotes the sampled noise. Sparse noise is denoted as a sparsity function is applied to add structural sparsity in the generated symmetric matrices. A random mask \( \mathbf{R} \in \mathbb{R}^{n \times n} \) is generated such that each element \( R_{ij} \) is sampled from a Bernoulli distribution with probability \( p = 0.5 \). The symmetric mask \( \mathbf{S} \) is then constructed as:
\begin{equation}
\mathbf{S} = \text{triu}(\mathbf{R}) + \text{triu}(\mathbf{R})^\top,
\end{equation}
where \( \text{triu}(\cdot) \) extracts the upper triangular portion of a matrix, including the diagonal, ensuring that \( \mathbf{S} \) is symmetric. The mask then applies element-wise, setting entries to zero wherever \( S_{ij} = 1 \):
\begin{equation}
\mathbf{M}_{\text{sparse}}[i, j] = 
\begin{cases} 
\mathbf{M}[i, j], & \text{if } S_{ij} = 0, \\ 
0, & \text{if } S_{ij} = 1.
\end{cases}
\end{equation}
These exploration strategies allow the critic to better understand the action space while maintaining reasonable preconditioners. Notably, any non-symmetric preconditioner is unsuitable, as the CG method requires symmetry for proper functionality.

Periodically, we sample \( \mathbf{A} \), \( \mathbf{M} \), and \( \mathbf{b} \) from the buffer to update both the actor \( \pi_\theta \) and the critic \( Q(\cdot) \), as defined in Equations \ref{eq:actor_loss} and \ref{eq:critic_loss}. Additionally, we ensure that the critic remains frozen during updates to the actor. To maintain stability, we clip the gradients of both the actor and the critic to a maximum norm, defined as a hyperparameter. Algorithm \ref{alg:train} demonstrates the PEARL training procedure.

\begin{algorithm}[h!]
\caption{PEARL Training Procedure}
\label{alg:train}
\small 
\begin{algorithmic}[1]
    \FOR{\texttt{epoch} from $1$ to \texttt{epochs}}
        \STATE $\gamma \leftarrow$ \texttt{condition\_scheduler.getValue()}
        \STATE $\epsilon \leftarrow$ \texttt{exploration\_scheduler.getValue()}
        \FOR{\texttt{step} from $1$ to \texttt{steps}}
            \STATE $\mathbf{A}, \mathbf{b} \leftarrow$ \texttt{generate\_system()}
            \STATE $\mathbf{M} \leftarrow \pi_\theta(\mathbf{A})$ 
            
            \COMMENT{Exploration}
            \STATE $p_1, p_2 \sim \text{Uniform}(0, 1)$
            \IF{$p_1 < \epsilon$}
                \STATE $\mathbf{M} \leftarrow \mathbf{M} + \mathbf{E}$, where $\mathbf{E} \sim \mathcal{N}(0, \sigma^2)$ and $\mathbf{E} = \mathbf{E}^T$
            \ENDIF
            \IF{$p_2 < \epsilon$}
                \STATE $\mathbf{M} \leftarrow \texttt{symmetric\_sparsify}(\mathbf{M})$
            \ENDIF
            \STATE \texttt{buffer.append($\mathbf{A}, \mathbf{b}, \mathbf{M}$)}
            \IF{$\texttt{step } \% \texttt{ wait} == 0$}
                \STATE $\mathbf{A}, \mathbf{b}, \mathbf{M} \leftarrow \texttt{buffer.sample()}$
                \STATE \COMMENT{Train Critic}
                \STATE $\mathbf{r} \leftarrow \mathcal{R}(\mathbf{A}, \mathbf{b}, \mathbf{M})$
                \STATE $\mathcal{L}_\phi \leftarrow  \left( \mathbf{r} - Q_\phi(\mathbf{A}_i, \mathbf{M}_i) \right)^2$
                \STATE $\mathbf{g}_\phi \leftarrow \nabla_\phi \mathcal{L}_\text{critic}$
                \STATE $\mathbf{g}_\phi \leftarrow \texttt{clip}(\mathbf{g}_\phi, \texttt{max\_norm})$
                \STATE $\phi \leftarrow \phi - \eta_\text{critic} \mathbf{g}_\phi$
                
                \COMMENT{Train Actor}
                \STATE \texttt{critic.freeze()}
                \STATE $\mathbf{M} \leftarrow \pi_\theta(\mathbf{A})$ 
                \STATE $\mathcal{L}_\theta \leftarrow  \gamma\log
                \left((
                    \log(\sigma_\text{max}) - \log(\sigma_\text{min}) \right) -Q_\phi(\mathbf{A}, \mathbf{M})$
                \STATE $\mathbf{g}_\theta \leftarrow \nabla_\theta \mathcal{L}_\theta$
                \STATE $\mathbf{g}_\theta \leftarrow \texttt{clip}(\mathbf{g}_\theta, \texttt{max\_norm})$
                \STATE $\theta \leftarrow \theta - \eta_\text{actor} \mathbf{g}_\theta$
            \ENDIF
        \ENDFOR
        \STATE \texttt{condition\_scheduler.step()}
        \STATE \texttt{exploration\_scheduler.step()}
    \ENDFOR
\end{algorithmic}
\end{algorithm}

\section{Theoretical Results}
In this section, we provide theoretical foundations to justify the assumptions underlying our numerical results. Specifically, we consider how preconditioning techniques achieve \(\epsilon\)-accurate solutions, and how to quantify an optimal preconditioner learning strategy. Detailed proofs for these results are provided in \ref{appendix:thr_res}.

Preconditioning forms the backbone of improving the convergence of iterative solvers. By introducing a preconditioner, the condition number of the system matrix can be significantly reduced, which impacts the convergence rate of various iterative solver algorithms. The following theorem formalizes this relationship, quantifying how the condition number of the preconditioned matrix influences the computational complexity of achieving an $\epsilon$-accurate solution using the CG method.

\begin{theorem} (Condition Number and CG Complexity) Let $\mathbf{A}$ be an $n \times n$ symmetric positive-definite (SPD) matrix, and let $\mathbf{M}$ be an $n \times n$ SPD preconditioner. Consider the preconditioned system

\begin{equation}
    \mathbf{M}^{-1} \mathbf{A} \mathbf{x}=\mathbf{M}^{-1} \mathbf{b}
\end{equation}
Define $\kappa=\kappa\left(\mathbf{M}^{-1} \mathbf{A}\right)=\frac{\lambda_{\max }\left(\mathbf{M}^{-1} \mathbf{A}\right)}{\lambda_{\min }\left(\mathbf{M}^{-1} \mathbf{A}\right)}$, where $\lambda_{\max }$ and $\lambda_{\text {min }}$ are the largest and smallest eigenvalues, respectively. The preconditioned CG algorithm applied to this system produces iterates $\mathbf{x}_k$ such that, for the error $\mathbf{e}_k=\mathbf{x}_k-\mathbf{x}^*$ (with $\mathbf{x}^*$ the true solution), the following bound holds:

\begin{equation}
    \left\|\mathbf{e}_k\right\|_{\mathbf{A}} \leq 2\left(\frac{\sqrt{\kappa}-1}{\sqrt{\kappa}+1}\right)^k\left\|\mathbf{e}_0\right\|_{\mathbf{A}}
\end{equation}
where $\|\mathbf{z}\|_{\mathbf{A}}=\sqrt{\mathbf{z}^{\top} \mathbf{A z}}$. Consequently, to achieve an $\epsilon$-accurate solution (e.g., to ensure $\left\|\mathbf{e}_k\right\|_{\mathbf{A}} \leq \epsilon\left\|\mathbf{e}_0\right\|_{\mathbf{A}}$ ), it suffices to take

\begin{equation}
k(\epsilon) \leq C \sqrt{\kappa} \log \left(\frac{1}{\epsilon}\right)
\end{equation}
for some universal constant $C$.
\end{theorem}

The effectiveness of a preconditioning strategy can be evaluated by its impact on the average condition number of a system, which influences the convergence of iterative solvers. By employing a learned policy that optimizes preconditioners, it is possible to systematically reduce the average condition number of systems drawn from a distribution. The following theorem establishes the improvement in CG complexity under such an optimized policy, providing a direct comparison with the case of no preconditioning.

\begin{theorem} (Improved Complexity under an Optimized Policy) Suppose the learned policy $\pi_\theta$ reduces the condition numbers of systems drawn from $D$ on average. Specifically, assume that for $s=(\mathbf{A}, \mathbf{b})$, the expected condition number after applying the learned preconditioner $\mathbf{M}_\theta(s)$ is $\mathbb{E}_{s \sim D}\left[\kappa\left(\mathbf{M}_\theta(s)^{-1} \mathbf{A}\right)\right] \leq \kappa_{\text {avg }}$. Then the expected number of CG iterations to achieve $\epsilon$-accuracy is:

\begin{equation}
\mathbb{E}_{s \sim D}[k(\epsilon)] \leq C \sqrt{\kappa_{\mathrm{avg}}} \log \left(\frac{1}{\epsilon}\right),
\end{equation}
which is less than or equal to the complexity under no preconditioning if $\kappa_{\text {avg }}<\kappa_{\text {avg-no-pre, }}$ the expected condition number without preconditioning.

\end{theorem}

\section{Numerical Results}

\begin{table*}[]
\centering
\resizebox{\textwidth}{!}{%
\begin{tabular}{lcc}
\toprule
\textbf{Preconditioner Model} & \textbf{Condition (Mean ± Std)} & \textbf{Iteration (Mean ± Std)} \\
\midrule
No Preconditioner & $5550.283 \pm 2418.825$ & $25.000 \pm 0.000$\\
Condition, Standard Training      & $12.389 \pm 4.265$ & $10.875 \pm 1.536$ \\
Critic, Cosine Schedule           & $(4.019 \pm 1.878) \times 10^{15}$ & $25.000 \pm 0.000$ \\
Condition/Critic, Fixed Schedule  & $12.083 \pm 4.713$ & $10.438 \pm 1.657$ \\
Condition/Critic, Cosine Schedule & $\mathbf{11.116} \pm \mathbf{4.794}$ & $\mathbf{10.000} \pm \mathbf{1.887}$ \\
\bottomrule
\end{tabular}%
}
\caption{Comparison of neural preconditioner training procedures. The best values in each column are highlighted in bold. While most condition-based methods achieve comparable results, the dual-objective cosine-scheduled procedure consistently yields the best (lowest) values.}
\label{tab:model_stats}
\end{table*}

\begin{figure*}[]
    \centering
    \begin{minipage}{0.33\textwidth}
        \centering
        \includegraphics[width=1.0\textwidth]{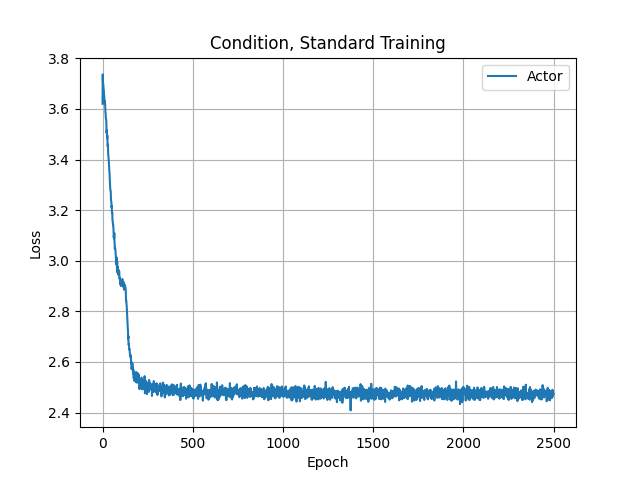}
    \end{minipage}\hfill
    \begin{minipage}{0.33\textwidth}
        \centering
        \includegraphics[width=1.0\textwidth]{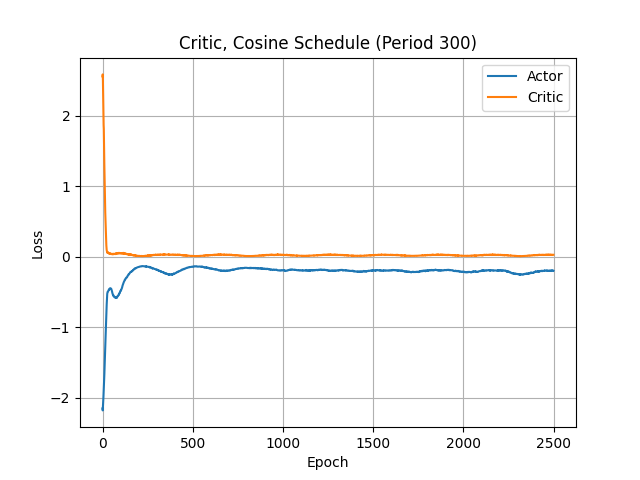}
    \end{minipage}
    \begin{minipage}{0.33\textwidth}
        \centering
        \includegraphics[width=1.0\linewidth]{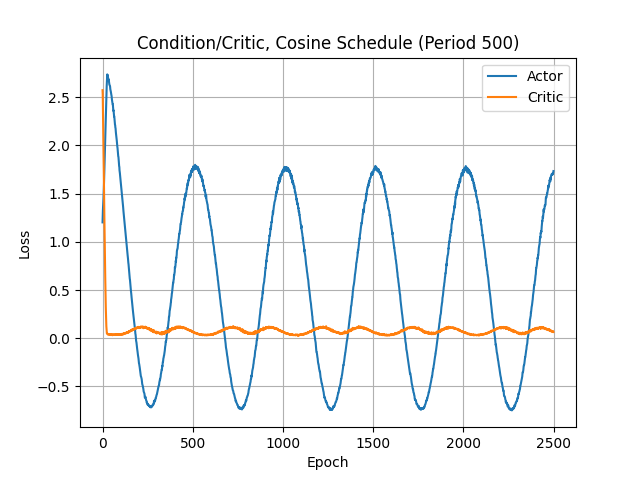}
    \end{minipage}
    \caption{Shown (left-to-right) are the training losses for the actor and critic using the standard procedure, cosine-scheduled procedure, and critic-only procedure. The theoretical minimum value of the critic loss is 0. For the actor loss, the theoretical minimum is -1.96, derived by summing the minimum condition effect (1.0) and the negated maximum actor reward -2.96.}
    \label{fig:training}
\end{figure*}

\begin{figure}[]
    \centering
    \includegraphics[width=0.55\linewidth]{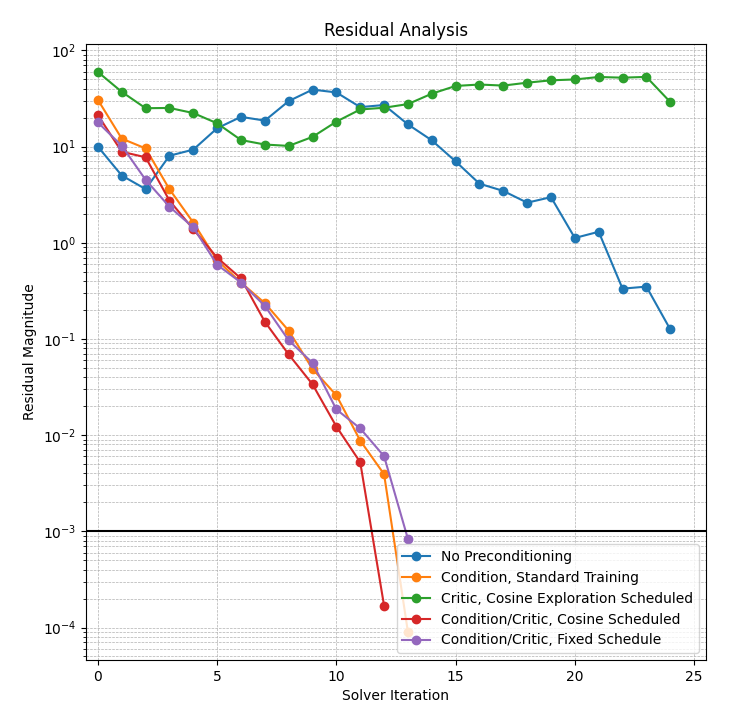}
    \caption{Residual analysis of the conjugate gradient (CG) solver was conducted on a randomly sampled system with an approximate condition number of 10,000. The results, depicted in the graph, demonstrate a slight advantage when employing the dual-objective cosine-scheduled training procedure.}
    \label{fig:residual-analysis}
\end{figure}

We present numerical results for five training procedures: no preconditioner, standard minimization of the condition number, maximization of the critic with a cosine schedule for exploration, dual-objective minimization with fixed exploration and condition effects, and dual-objective minimization with a cosine schedule applied to both exploration and condition effects, as summarized in Table \ref{tab:model_stats}. In addition, we provide a residual analysis of each method on a randomly sampled system, shown in Figure \ref{fig:residual-analysis}. Training losses for the three key techniques—condition-only, critic-only, and combined condition/critic—are shown in Figure \ref{fig:training}. Additionally, we compare these results with the traditional preconditioners Jacobi, ILU, and AMG, whose effectiveness is further detailed in \ref{appendix:C}, Figure \ref{fig:trad}. We do not included results comparing multi-reward and single-reward critics as we found the differences between the systems to insignificant, therefore all models (that use a critic) use a single-reward critic. 

Quantitatively, we observe that condition-based models achieve approximately a 60\% improvement, outperforming ILU with fill-level 1 but falling short of AMG with a maximum level of 1. As shown in Figure \ref{fig:training}, the condition-only model rapidly minimizes the condition number but eventually oscillates, making it susceptible to getting trapped in local minima. In contrast, condition models augmented with a critic—particularly those employing a cosine schedule—undergo phases of high exploration. While these phases often yield suboptimal preconditioners, they expand the minimization problem to other structurally plausible regions of the search space. This observation is further supported by the performance of the condition/critic model with a fixed schedule, which avoids high exploration phases and behaves similarly to standard condition number minimization. Our proposed training procedure, leveraging the dual-objective framework with a cosine schedule, achieves approximately a 4\% improvement over the already competitive condition number minimization approach.

\begin{figure*}
    \begin{minipage}{1.0\textwidth}
        \centering
        \includegraphics[width=0.6\linewidth]{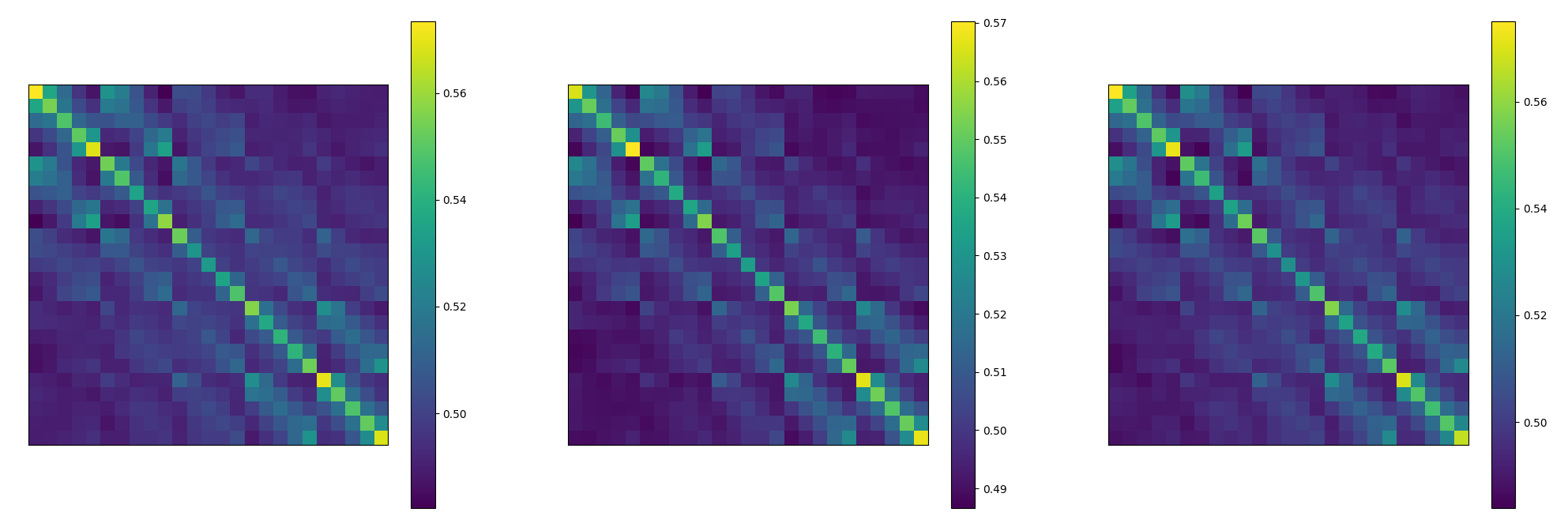}
    \end{minipage}\hfill
    \begin{minipage}{1.0\textwidth}
        \centering
        \includegraphics[width=0.6\linewidth]{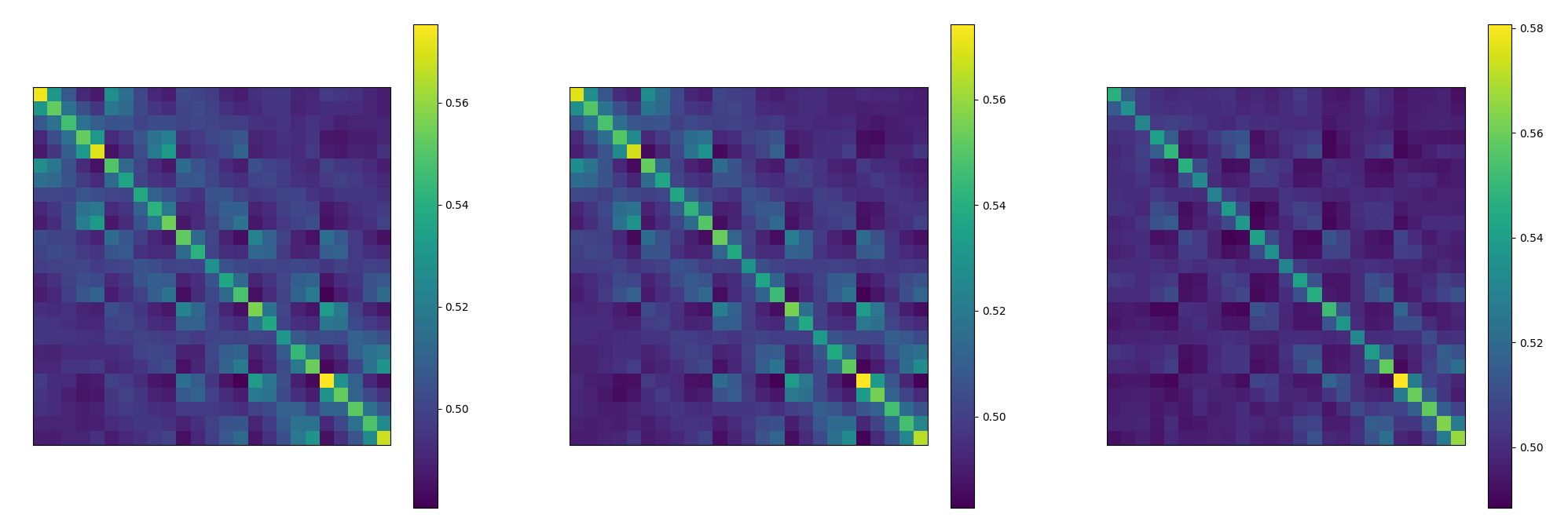}
    \end{minipage}\hfill
    \begin{minipage}{1.0\textwidth}
        \centering
        \includegraphics[width=0.6\linewidth]{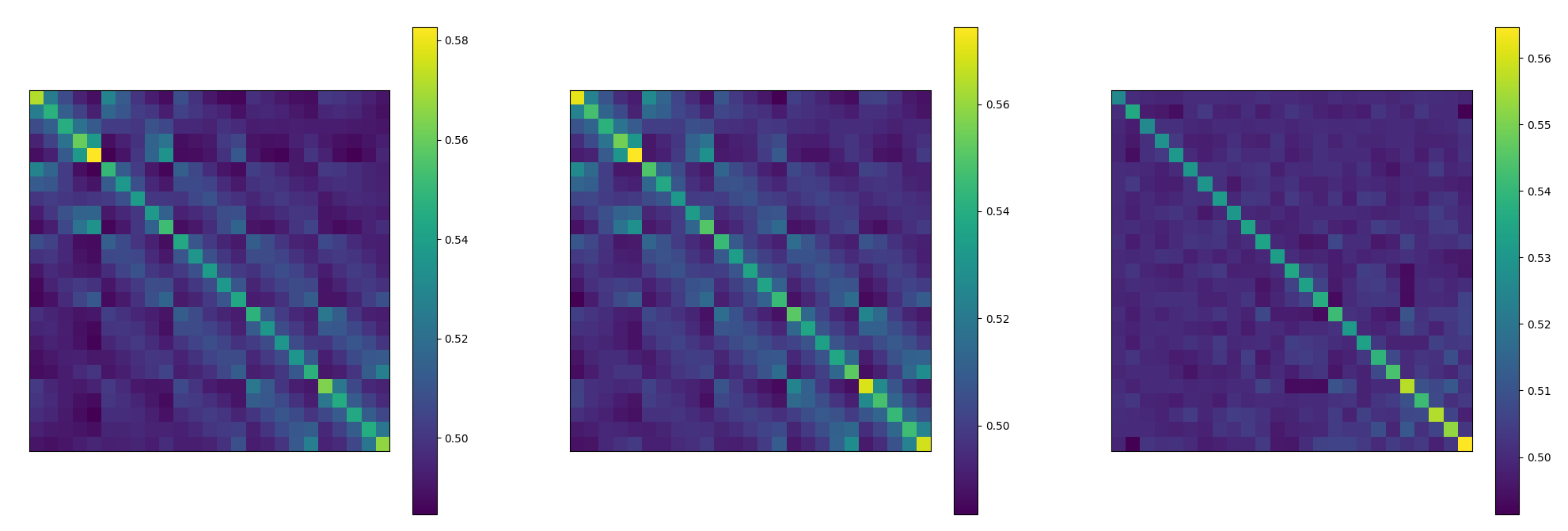}
    \end{minipage}\hfill
    \begin{minipage}{1.0\textwidth}
        \centering
        \includegraphics[width=0.6\linewidth]{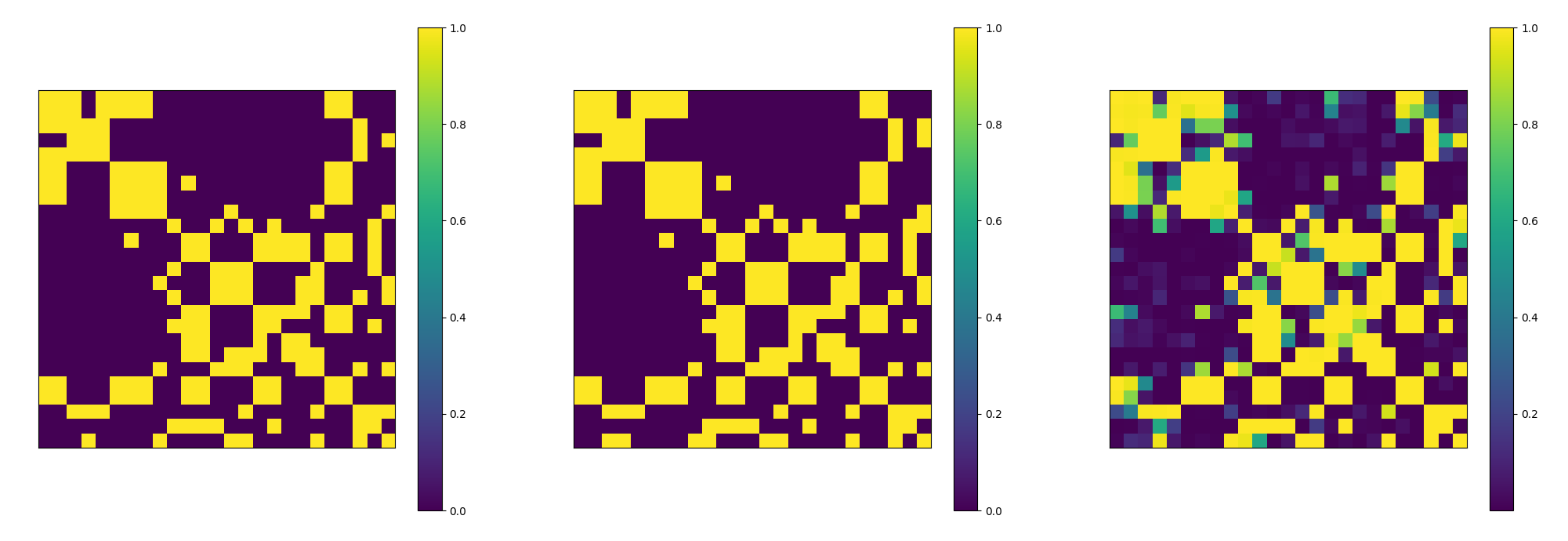}
    \end{minipage}\hfill
    \caption{Shown are the timesteps at 2500, 1000, and 100 for the four models compared in Table \ref{tab:model_stats}. From top to bottom, the models are condition-only, condition/critic fixed schedule, condition/critic cosine-scheduled, and critic-only. Qualitatively, the condition/critic cosine-scheduled model produces more organized preconditioners, but takes longer to converge. Additionally, the critic-only model exhibits slow structural changes, where only one or two elements of the preconditioner are adjusted at a time.}
    \label{fig:extra_out}
\end{figure*}

Qualitatively, we observe that condition/critic models produce preconditioners that are structurally more robust and resemble the patterns seen in AMG with a maximum level of 1. Comparing Figures \ref{fig:cond-v-cond_critic} and \ref{fig:trad}, both display a structure similar to AMG; however, the condition-only model exhibits significantly more noise, resulting in a less organized preconditioned system. Conversely, the condition/critic model with a cosine-scheduled training procedure generates a preconditioner that is more organized and less noisy, aligning more closely with the structure of AMG preconditioners. Figure \ref{fig:extra_out} illustrates the learned preconditioner at different timesteps: 2500, 1000, and 100 (from left to right). At the top, the condition-only model quickly converges to a noisy pattern within the first 100 iterations. The condition/critic model with a fixed schedule (second from the top) also quickly learns a noisy pattern. In contrast, the condition/critic model with a cosine schedule takes significantly longer to converge but ultimately forms a structure much closer to what we expect for effective preconditioning. Another qualitative observation is the presence of inverse diagonals in the preconditioners. Ideally, all valid preconditioners should include the inverse diagonals of the system matrix. However, only the condition/critic model with a cosine schedule achieves this property, while other models fail to consistently approximate the diagonals, often producing nearly uniform values along each forward pass.

We also present t-SNE plots of various training procedures in \ref{appendix:C}, Figure \ref{fig:exploration}, under different levels of exploration. Our results show that as the cosine period increases, the separation within the latent space becomes more pronounced, providing a clear indication of additional exploration. Condition-only models exhibit a highly singular cluster with minimal disjoint regions, while critic-only models display extreme separation, forming multiple distinct clusters. Interestingly, applying cosine-scheduled exploration in the critic-only setting significantly smooths the exploration process, addressing the stability issues observed in the fixed-schedule case. By combining condition and critic objectives, we achieve a unified framework that allows control over the level of exploration by adjusting the cosine period length. Specifically, longer periods result in improved exploration. The best-performing model from Table \ref{tab:model_stats} used a period of 500, the largest value we experimented with.

\section{Discussion}

We find that our approach produces both superior quantitative and qualitative results compared to standard condition-only training procedures. Specifically, our framework achieves a 4\% improvement in solver time and a corresponding reduction in the condition number, while also generating significantly more organized preconditioners. Through t-SNE component plots, we demonstrate that our method’s exploration can be effectively controlled by adjusting the length of the cosine-scheduled period. Longer periods promote greater exploration and better convergence. However, this balance is delicate: excessive exploration may hinder the exploitation of new knowledge, while insufficient exploration can lead to suboptimal results, as evidenced by the condition-only model.

Using this framework to learn the ILU decomposition instead of the incomplete Cholesky decomposition, we observe that the model significantly reduces the condition number but fails to accelerate the solver. This outcome arises because the LU decomposition does not inherently preserve the symmetric positive definite (SPD) property. Consequently, the condition gradient becomes unstable, resulting in suboptimal solutions, as illustrated in Figure \ref{fig:lu-cond}. By incorporating the critic into the optimization process, we mitigate collapse into suboptimal solutions; however, we still encounter challenges in fitting an effective preconditioner. This observation leads us to hypothesize that the condition number primarily guides the actor toward the optimal solution, while the critic facilitates structured exploration of the search space through gradual, incremental updates. This finding is significant as it suggests that this reinforcement learning framework can be further analyzed to quantify the balance between exploitation and exploration in the learned preconditioner policies. Furthermore, we provide outputs from time steps 2500, 1000, and 100 of the critic-only cosine-scheduled model, shown at the bottom of Figure \ref{fig:extra_out}.

\noindent{\textbf{Limitations.}} While the proposed framework is evaluated on 2D diffusion equations with random perturbations—thereby capturing a class of ill-conditioned systems—its performance on more diverse PDEs, particularly those involving nonlinearities, complex boundary conditions, or domain-specific structures, remains untested. Furthermore, although the experimental setup includes moderate-size matrices, large-scale scientific applications involve systems with millions of degrees of freedom. Training an actor-critic agent on matrices of this size may require extensive memory, complex sampling strategies, and distributed computing infrastructures. Assessing whether the method can be feasibly scaled up without sacrificing accuracy or incurring prohibitive computational costs is an open question. In addition, the training procedure uses both condition-number gradients and rewards derived from convergence metrics, each of which can be numerically delicate. Condition-number calculations can become unstable for matrices approaching singularity, leading to noisy gradient signals. Although the introduction of a cosine scheduler helps moderate these effects, the stability of the training loop still depends on heuristics such as scheduler periods, gradient clipping thresholds, and loss-function weights. An inadequate choice of hyperparameters risks suboptimal convergence or unstable training dynamics.

\section{Conclusion and Future Work}

In this paper, we present an approach to generalizing preconditioner learning. We propose a dual-objective framework that captures key characteristics of effective preconditioners, including minimizing solver runtime and improving the condition number. We demonstrate that this framework performs comparably to traditional preconditioners and state-of-the-art deep learning methods, while also exploring a broader preconditioner space and mitigating catastrophic convergence issues. Additionally, we introduce the use of a cosine scheduler to balance exploration, the condition number, and manage the buffer. Finally, we hypothesize potential limitations of the critic-based approach in this context.

We propose several key extensions to this framework. First, we suggest using a multistep Markov decision process (MDP) that operates on residual vectors rather than the system itself. This extension would require target networks and could stabilize the critic's updates, though this stability comes at the cost of removing the condition number from consideration. The reward at each step could be based on the magnitude of the residual reduction, encouraging large, rapid decreases in the residual vector during each iteration. Additionally, we propose an intriguing direction as extending this framework to other iterative solvers. In this study, we focused exclusively on SPD systems, which motivated our use of the CG method. However, solvers such as GMRES \cite{doi:10.1137/0907058} and MINRES \cite{VanDerVorst2003}, which are well-suited for non-symmetric and indefinite systems, could benefit from this framework's additional stabilization. Another promising area of exploration is applying this framework to higher-dimensional systems, particularly with the integration of more scalable models. As previously discussed, CNNs are often the default choice for high-dimensional problems but tend to struggle with sparsity. Sparse convolutions, on the other hand, may offer a scalable solution with minimal impact on the model's ability to learn, potentially unlocking new possibilities for efficiently handling high-dimensional, sparse data.


 \bibliographystyle{elsarticle-num} 
 \bibliography{ref}


\newpage
\appendix
\section{Theoretical Result Proofs}\label{appendix:thr_res}

\begin{proof}[Proof for Theorem 1] \label{proof:1}
Consider the linear system $\mathbf{A x}=\mathbf{b}$ and its preconditioned form $\mathbf{M}^{-1} \mathbf{A} \mathbf{x}=\mathbf{M}^{-1} \mathbf{b}$. Let $\mathbf{x}^*$ be the unique solution. The CG method applied to the preconditioned system generates a sequence of approximations $\mathbf{x}_k$ in the Krylov subspace:

\begin{equation}
\mathcal{K}_k\left(\mathbf{M}^{-1} \mathbf{A}, \mathbf{r}_0\right) = 
\operatorname{span}\left\{
\mathbf{r}_0, 
\mathbf{M}^{-1} \mathbf{A} \mathbf{r}_0, 
\left(\mathbf{M}^{-1} \mathbf{A}\right)^2 \mathbf{r}_0, 
\ldots, 
\left(\mathbf{M}^{-1} \mathbf{A}\right)^{k-1} \mathbf{r}_0
\right\}.
\end{equation}
where $\mathbf{r}_0=\mathbf{b}-\mathbf{A} \mathbf{x}_0$ is the initial residual (assuming some initial guess $\mathbf{x}_0$ ). Define the energy norm $\|\mathbf{z}\|_{\mathbf{A}}=\sqrt{\mathbf{z}^{\top} \mathbf{A z}}$. The CG method, at the $k$-th iteration, chooses $\mathbf{x}_k \in \mathbf{x}_0+\mathcal{K}_k\left(\mathbf{M}^{-1} \mathbf{A}, \mathbf{r}_0\right)$ to minimize the A-norm of the error:

\begin{equation}
\mathbf{x}_k=\arg \min _{\mathbf{x}_0+\mathcal{K}_k}\left\|\mathbf{x}-\mathbf{x}^*\right\|_{\mathbf{A}} .
\end{equation}
Equivalently, $\mathbf{r}_k=\mathbf{b}-\mathbf{A} \mathbf{x}_k$ is the residual that is orthogonal (in the $\mathbf{M}^{-1}$-weighted inner product) to the subspace $\mathcal{K}_k$. A classical approach to bounding the CG error involves polynomial approximation. Let $\lambda_1 \leq \lambda_2 \leq \cdots \leq \lambda_n$ be the eigenvalues of $\mathbf{M}^{-1} \mathbf{A}$. By definition, $\kappa=\lambda_n / \lambda_1$.
Consider the set $\Lambda=\left[\lambda_1, \lambda_n\right]$ containing all eigenvalues. The CG method's error bound can be understood by examining the error projection onto the eigenbasis of $\mathbf{M}^{-1} \mathbf{A}$. Specifically, after $k$ steps, the CG error is bounded by the best approximation of zero by polynomials $p$ of degree $k$ that satisfy $p(0)=1$, applied to $\mathbf{M}^{-1} \mathbf{A}$:

\begin{equation}
\left\|\mathbf{e}_k\right\|_{\mathbf{A}}=\min _{p \in \mathcal{P}_k, p(0)=1}\left\|p\left(\mathbf{M}^{-1} \mathbf{A}\right) \mathbf{e}_0\right\|_{\mathbf{A}}
\end{equation}
Since $\mathbf{e}_0$ can be expanded in the eigenbasis, it suffices to control $\max _{\lambda \in \Lambda}|p(\lambda)|$. It is known from approximation theory that the polynomial which minimizes the maximum error on the interval $\left[\lambda_1, \lambda_n\right]$ subject to $p(0)=1$ can be constructed by scaling and shifting the Chebyshev polynomials of the first kind. The key known result is:

\begin{equation}
\min _{p \in \mathcal{P}_k, p(0)=1} \max _{\lambda \in\left[\lambda_1, \lambda_n\right]}|p(\lambda)| \leq 2\left(\frac{\sqrt{\lambda_n}-\sqrt{\lambda_1}}{\sqrt{\lambda_n}+\sqrt{\lambda_1}}\right)^k
\end{equation}
This result comes from the fact that Chebyshev polynomials $T_k(x)$ have minimal uniform deviation from zero on $[-1,1]$. By using the linear mapping that takes $\lambda_1$ to 1 and $\lambda_n$ to $\kappa$, and then a further transform to map $\left[\lambda_1, \lambda_n\right]$ into a symmetric interval about zero, one obtains the above sharp bound. The factor 2 emerges from normalization conditions. 
Substitute $\lambda_{\min }=\lambda_1, \lambda_{\max }=\lambda_n:$ Rewriting the bound in terms of $\kappa=\lambda_{\max } / \lambda_{\min }$, we get:

\begin{equation}
\left\|\mathbf{e}_k\right\|_{\mathbf{A}} \leq 2\left(\frac{\sqrt{\kappa}-1}{\sqrt{\kappa}+1}\right)^k\left\|\mathbf{e}_0\right\|_{\mathbf{A}}
\end{equation}
To achieve an accuracy $\left\|\mathbf{e}_k\right\|_{\mathbf{A}} \leq \epsilon\left\|\mathbf{e}_0\right\|_{\mathbf{A}}$, we require:

\begin{equation}
2\left(\frac{\sqrt{\kappa}-1}{\sqrt{\kappa}+1}\right)^k \leq \epsilon
\end{equation}
Taking logarithms
$
\log \left(\frac{2}{\epsilon}\right) \geq k \log \left(\frac{\sqrt{\kappa}+1}{\sqrt{\kappa}-1}\right).
$
Notice that:
$
\frac{\sqrt{\kappa}+1}{\sqrt{\kappa}-1}=1+\frac{2}{\sqrt{\kappa}-1} .
$
For large $\kappa, \frac{\sqrt{\kappa}+1}{\sqrt{\kappa}-1} \approx \sqrt{\kappa}$. More precisely, $\log \left(\frac{\sqrt{\kappa}+1}{\sqrt{\kappa}-1}\right)$ behaves like $\log \left(\kappa^{1 / 2}\right)=\frac{1}{2} \log (\kappa)$ for large $\kappa$.
Thus, asymptotically:

\begin{equation}
k \geq \frac{\log (2 / \epsilon)}{\log \left(\frac{\sqrt{\kappa}+1}{\sqrt{\kappa}-1}\right)}
\end{equation}
Since $\log \left(\frac{\sqrt{\kappa}+1}{\sqrt{\kappa}-1}\right) \approx \log (\sqrt{\kappa})=\frac{1}{2} \log (\kappa)$, we get:

\begin{equation}
k(\epsilon)=O(\sqrt{\kappa} \log (1 / \epsilon)) .
\end{equation}
By adjusting constants, one can write:

\begin{equation}
k(\epsilon) \leq C \sqrt{\kappa} \log \left(\frac{1}{\epsilon}\right)
\end{equation}
for some universal constant $C$.

\end{proof}

\begin{proof}[Proof for Theorem 2]\label{proof:2}
From Theorem 1, for a single fixed system $\mathbf{A} \boldsymbol{x}=\mathbf{b}$ and a given SPD preconditioner $\mathbf{M}$, we know that the CG method applied to $\mathbf{M}^{-1} \mathbf{A} \boldsymbol{x}=\mathbf{M}^{-1} \mathbf{b}$ converges at a rate governed by $\kappa=\kappa\left(\mathbf{M}^{-1} \mathbf{A}\right)$. Specifically, to achieve a relative error $\left\|\mathbf{x}_k-\mathbf{x}^*\right\|_{\mathbf{A}} \leq \epsilon\left\|\mathbf{x}_0-\mathbf{x}^*\right\|_{\mathbf{A}}$, the number of iterations $k(\epsilon)$ satisfies:

\begin{equation}
k(\epsilon) \leq C \sqrt{\kappa} \log \left(\frac{1}{\epsilon}\right),
\end{equation}
where $C$ is a universal constant. Applying the Learned Policy: Now consider a distribution $D$ over states $s=(\mathbf{A}, \mathbf{b})$. The learned policy $\pi_\theta$ assigns a preconditioner $\mathbf{M}_\theta(s)$ to each state. For each $s$, define $\kappa_s=$ $\kappa\left(\mathbf{M}_\theta(s)^{-1} \mathbf{A}\right)$. Applying the same complexity bound from Theorem 1 to each system $s$ independently, we get 
$
k_s(\epsilon) \leq C \sqrt{\kappa_s} \log \left(\frac{1}{\epsilon}\right) .
$
Now consider a distribution $D$ over states $s=(\mathbf{A}, \mathbf{b})$. The learned policy $\pi_\theta$ assigns a preconditioner $\mathbf{M}_\theta(s)$ to each state. For each $s$, define $\kappa_s=$ $\kappa\left(\mathbf{M}_\theta(s)^{-1} \mathbf{A}\right)$. Applying the same complexity bound from Theorem 1 to each system $s$ independently, we get:

\begin{equation}
k_s(\epsilon) \leq C \sqrt{\kappa_s} \log \left(\frac{1}{\epsilon}\right) .
\end{equation}
Factor out constants and terms not depending on $s$ :

\begin{equation}
\mathbb{E}_{s \sim D}[k(\epsilon)] \leq C \log \left(\frac{1}{\epsilon}\right) \mathbb{E}_{s \sim D}\left[\sqrt{\kappa_s}\right] .
\end{equation}
Rearranging
$
\mathbb{E}\left[\sqrt{\kappa_s}\right] \leq \sqrt{\mathbb{E}\left[\kappa_s\right]}.
$
Thus,
$
\mathbb{E}\left[\sqrt{\kappa_s}\right] \leq \sqrt{\kappa_{\mathrm{avg}}}.
$
Substitute $\mathbb{E}\left[\sqrt{\kappa_s}\right] \leq \sqrt{\kappa_{\text {avg }}}$ back into the expectation bound:

\begin{equation}
\mathbb{E}_{s \sim D}[k(\epsilon)] \leq C \log \left(\frac{1}{\epsilon}\right) \sqrt{\kappa_{\text {avg }}}.
\end{equation}
This shows that the expected iteration count for CG under the learned preconditioner is governed by the square root of the expected condition number $\kappa_{\text {avg }}$. Without any preconditioning, the expected condition number might be $\kappa_{\text {avg-no-pre, }}$ potentially much larger than $\kappa_{\text {avg. }}$. Since the complexity scales as $\sqrt{\kappa_{\text {avg-no-pre, }}}$ if the learned policy yields $\kappa_{\text {avg }}<\kappa_{\text {avg-no-pre, }}$ we have:

\begin{equation}
\mathbb{E}_{s \sim D}[k(\epsilon)] \leq C \sqrt{\kappa_{\text {avg }}} \log \left(\frac{1}{\epsilon}\right)<C \sqrt{\kappa_{\text {avg-no-pre }}} \log \left(\frac{1}{\epsilon}\right)
\end{equation}
thus confirming an improvement in expected complexity. In conclusion, under some assumptions, the learned policy's ability to reduce the expected condition number $\mathbb{E}\left[\kappa_s\right]$ directly translates into a lower expected iteration count for CG. The linearity of expectation and Jensen's inequality provide the link from individual problem complexity bounds to expected complexity under distribution $D$. 

\end{proof}


%


\section{Code Availability}\label{appendix:A}
Our repository is available under
the Apache License version 2.0 at
\url{https://github.com/djm3622/precondition-discovery-contextual-bandit}.

\section{Additional Actors}
\label{appendix:actors}
This section describes the additional actors included in our repository. We provide two models that preserve symmetry (including the incomplete Cholesky decomposition) and two that do not. Among these, the incomplete Cholesky decomposition model is the only one that also ensures positive definiteness.

\subsection*{Actor (ILU Decomposition)}
 Given an input system matrix $\mathbf{A} \in \mathbb{R}^{n \times n}$, the model predicts a lower triangular matrix $\mathbf{L} \in \mathbb{R}^{n \times n}$ and an upper triangular matrix $\mathbf{U} \in \mathbb{R}^{n \times n}$ such that:
\begin{equation}
    \mathbf{M} = \mathbf{L} \mathbf{U}.
\end{equation}
The model is parameterized as a fully connected neural network with configurable hidden sizes and a variable number of layers. At the midpoint of the network, we introduce a shared layer that branches into two separate sub-networks. These sub-networks are designed to output the lower triangular elements of $\mathbf{L}$ and the upper triangular elements of $\mathbf{U}$, respectively, corresponding to the indices specified by:
\begin{align}
\text{tril}(\mathbf{L}) &= \{\mathbf{L}_{ij} \mid i \geq j\}, \notag\\
\text{tril}(\mathbf{U}) &= \{\mathbf{U}_{ij} \mid i \leq j\}.
\end{align}
We then enforce sparsity and promote the SPD property in the same manner as the incomplete Cholesky decomposition model, following Equations \ref{eq:sparse} and \ref{eq:positive}. However, this model does not yield a stable estimate for the condition number.

\subsection*{Actor (Additive Symmetric)}
 Given an input system matrix $\mathbf{A} \in \mathbb{R}^{n \times n}$, the model predicts a lower triangular matrix $\mathbf{L} \in \mathbb{R}^{n \times n}$ such that:
\begin{equation}
    \mathbf{M} = \mathbf{L} + \mathbf{L}^\top - \text{diag}(\mathbf{L}).
\end{equation}
The model is parameterized similarly to the incomplete Cholesky decomposition model. It enforces sparsity and promotes the SPD property using Equations \ref{eq:sparse} and \ref{eq:positive}. The results obtained with this model are comparable to those achieved with the incomplete Cholesky decomposition approach.

\subsection*{Actor (Full Matrix)}
Given an input system matrix $\mathbf{A} \in \mathbb{R}^{n \times n}$, the model predicts the full preconditioner matrix $\mathbf{M} \in \mathbb{R}^{n \times n}$. The model is parameterized by a large fully connected network. Sparsity and the SPD property are encouraged in the same manner as the incomplete Cholesky decomposition model, using Equations \ref{eq:sparse} and \ref{eq:positive}. Among the four models we experimented with, this model proved to be the most unstable, and we do not have any significant results to report.

\section{Additional Figures}\label{appendix:C}
In this section, we present the outputs from the incomplete LU decomposition models and traditional preconditioners. Additionally, we provide a qualitative comparison between the condition-only and condition/critic cosine-scheduled models, t-SNE component plots, and various outputs from the four models compared in Table \ref{tab:model_stats}.

\begin{figure}[h]
    \begin{minipage}
        {0.5\textwidth}
        \centering
        \includegraphics[width=0.8\linewidth]{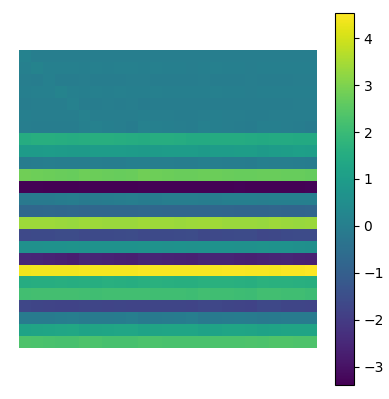}
        \caption{Output from a condition-only logit model, estimating the incomplete LU factorization. The model collapses to a completely collinear matrix. The condition number reaches nearly 1.0 but the solver runtime is worsened.}
        \label{fig:lu-cond}
    \end{minipage}
    \begin{minipage}
        {0.5\textwidth}
        \centering
        \includegraphics[width=0.8\linewidth]{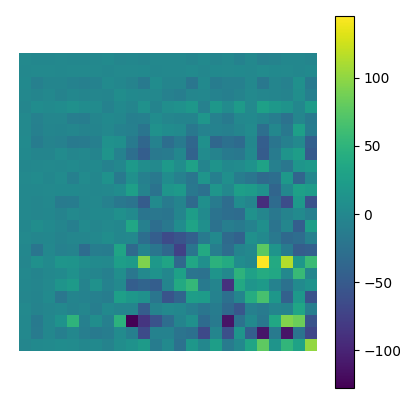}
        \caption{Output from a dual-optimization logit model, estimating the incomplete LU factorization, the model does not collapse the same as before but it struggles to learn much from the unstable critic updates.}
    \end{minipage}
\end{figure}

\begin{figure*}
    \centering
    \includegraphics[width=1.0\linewidth]{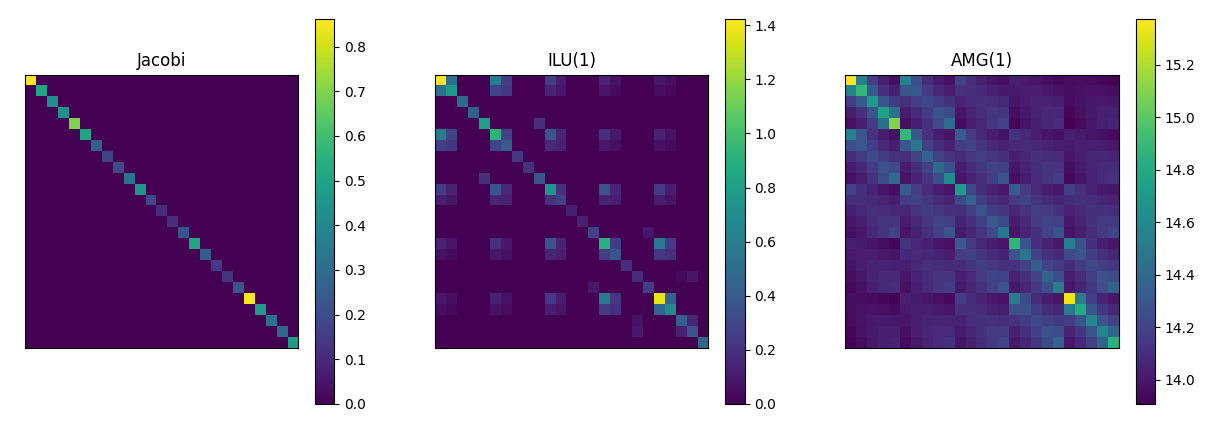}
    \caption{Traditional preconditioners. Shown from left to right are the Jacobi, ILU with fill-level one, and classical AMG with a maximum-level one. The Jacobi method results in no speedup (no convergence), the ILU method achieves a $20\%$ speedup (convergence), and the AMG method provides a $100\%$ speedup (via approximation of the inverse).}
    \label{fig:trad}
\end{figure*}

\begin{figure*}
    \begin{minipage}
        {1.0\textwidth}
        \centering
        \includegraphics[width=1.0\linewidth]{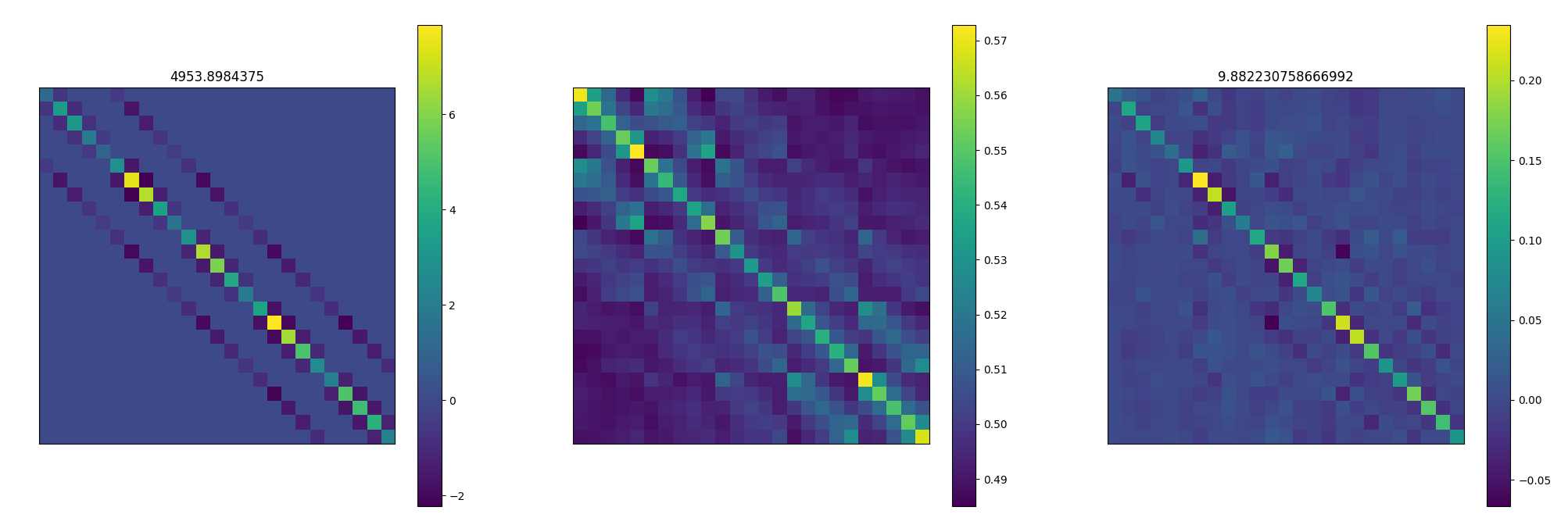}
    \end{minipage}
    \begin{minipage}
        {1.0\textwidth}
        \centering
        \includegraphics[width=1.0\linewidth]{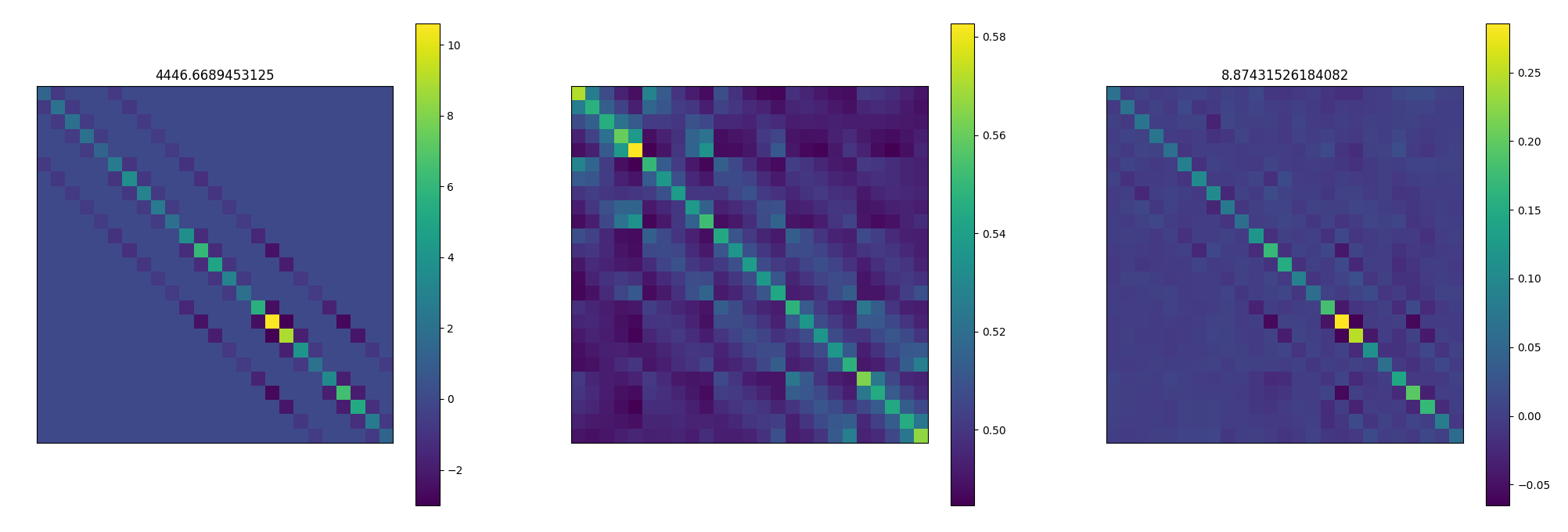}
    \end{minipage}
    \caption{Pictured on the top is the results for the condition-only model, while at the bottom, we is the results for the condition/critic model. The condition/critic model produces preconditioners that adhere much more closely to the expected structural patterns, resembling the AMG method. Additionally, the preconditioned system resulting from the condition/critic model is significantly more organized, highlighting its effectiveness in creating structured preconditioners.}
    \label{fig:cond-v-cond_critic}
\end{figure*}

\begin{figure*}
    \begin{minipage}
        {0.5\textwidth}
        \centering
        \includegraphics[width=1.0\linewidth]{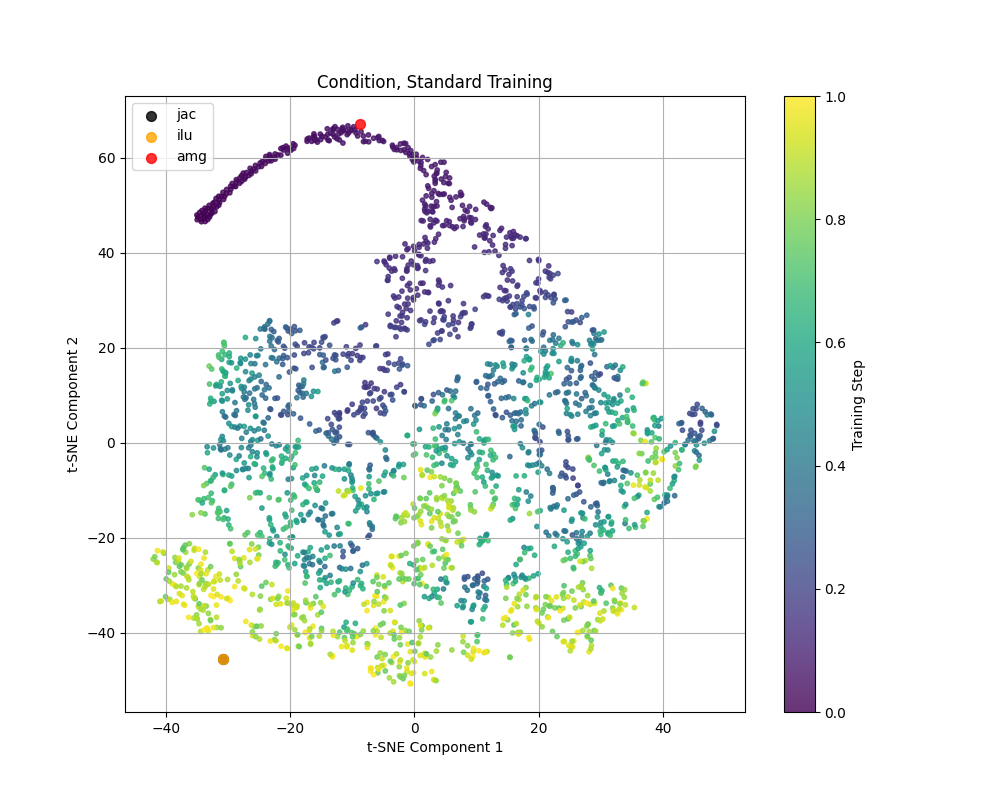}
    \end{minipage}
    \begin{minipage}
        {0.5\textwidth}
        \centering
        \includegraphics[width=1.0\linewidth]{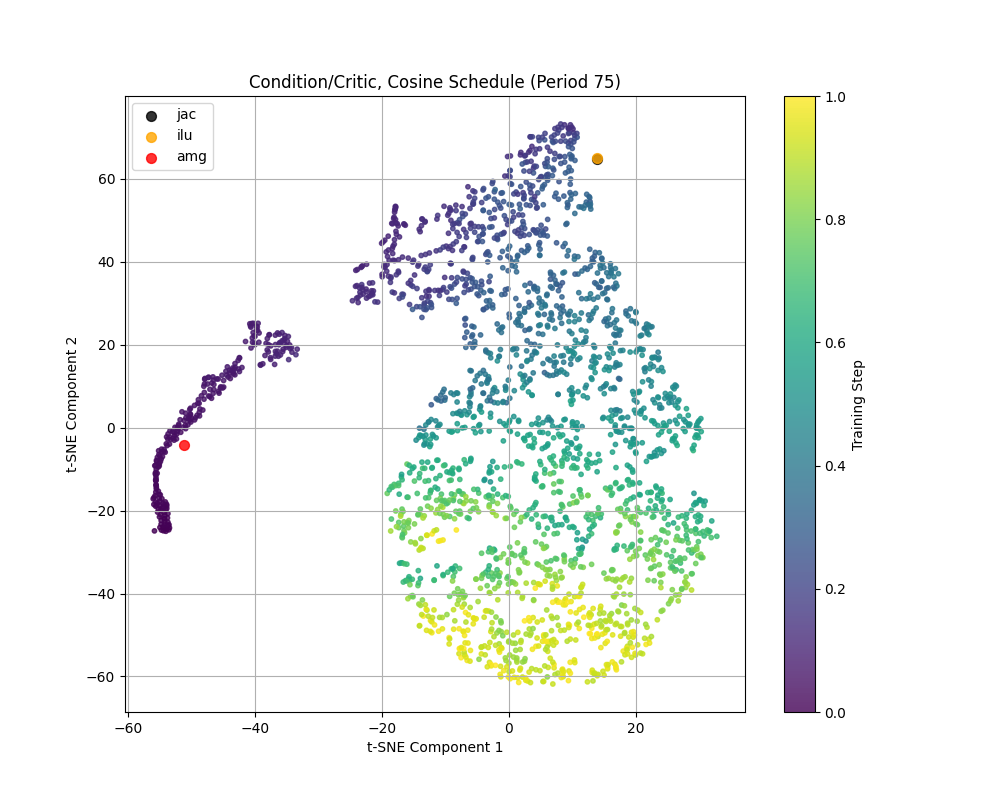}
    \end{minipage}
    \begin{minipage}
        {0.5\textwidth}
        \centering
        \includegraphics[width=1.0\linewidth]{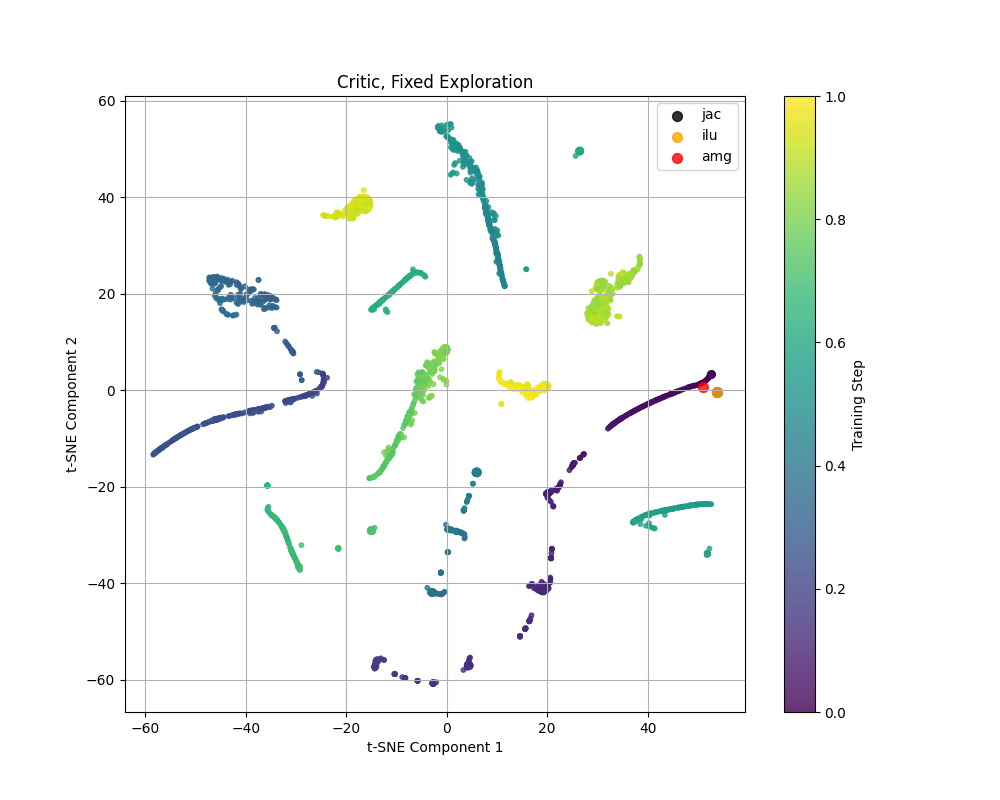}
    \end{minipage}
    \begin{minipage}
        {0.5\textwidth}
        \centering
        \includegraphics[width=1.0\linewidth]{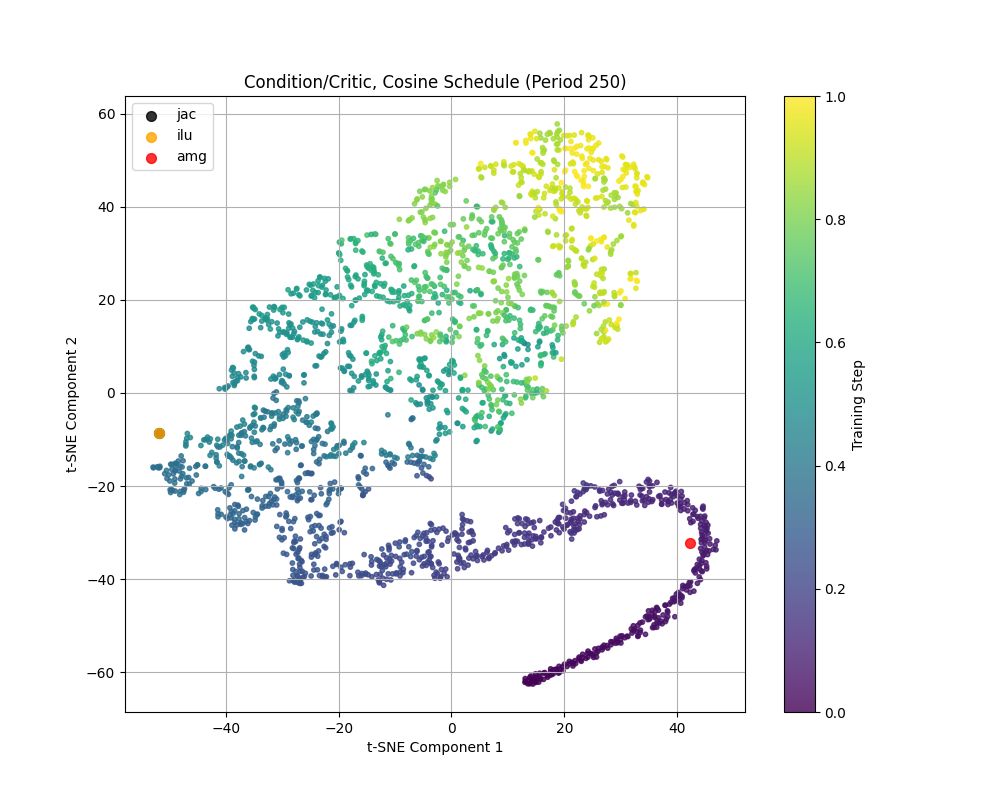}
    \end{minipage}
    \begin{minipage}
        {0.5\textwidth}
        \centering
        \includegraphics[width=1.0\linewidth]{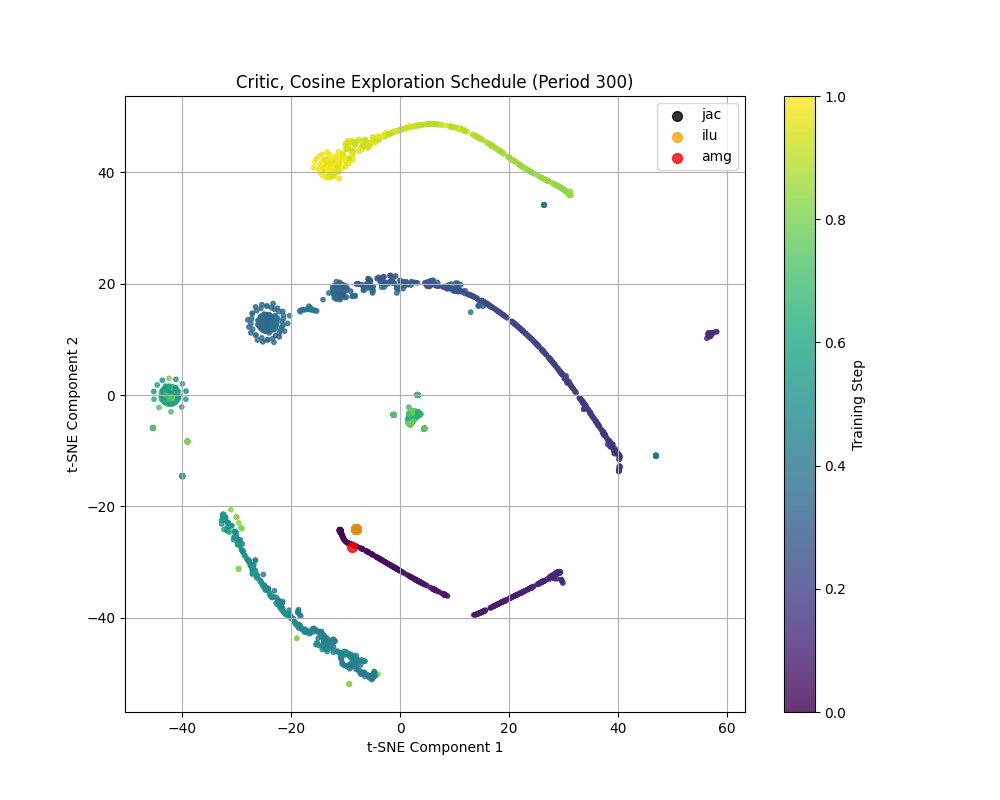}
    \end{minipage}
    \begin{minipage}
        {0.5\textwidth}
        \centering
        \includegraphics[width=1.0\linewidth]{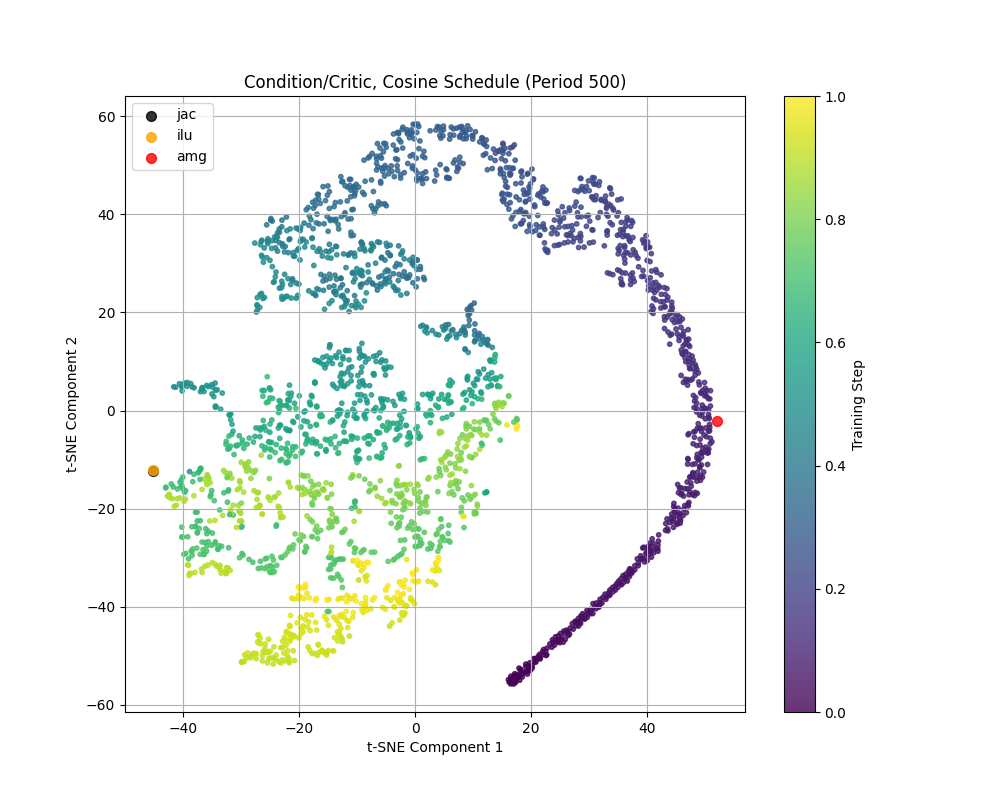}
    \end{minipage}
    \caption{Presented are the t-SNE component plots of the following models (arranged from left-to-right, top-to-bottom): condition-only, condition/critic cosine-scheduled with a period of 75, critic-only fixed-schedule, condition/critic cosine-scheduled with a period of 250, critic-only cosine-scheduled, and condition/critic cosine-scheduled with a period of 500. These plots demonstrate that our PEARL framework effectively combines the structured nature of the condition model with the inherent exploratory capabilities of the critic model. The degree of exploration can be controlled by adjusting the length of the cosine-scheduled period.}
    \label{fig:exploration}
\end{figure*}

\end{document}